\documentclass[manuscript,screen,nonacm]{acmart}
\setcopyright{none}
\usepackage{booktabs}
\usepackage{graphicx}
\usepackage{float}
\usepackage{adjustbox}

\makeatletter\def\@ACM@checkaffil{}\makeatother
\begin{document}

\title{Same Person, Different Advice: Auditing LLM Financial Advice with Repeated Risk-Tolerance Responses}

\author{Soovadeep Bakshi}
\email{soovadeep.bakshi@utexas.edu}
\affiliation{\institution{Independent Researcher}}
\author{Apurba Bose}
\affiliation{\institution{Independent Researcher}}
\author{Aishwarya Bose}
\affiliation{\institution{Independent Researcher}}
\author{Sandip Nandi}
\affiliation{\institution{Independent Researcher}}

\begin{abstract}
AI financial advisers set a client's stock allocation largely from one stated-risk question, yet people do not
answer it the same way twice: in a national US panel, 41\% of households chose a different answer two years later,
and in a laboratory retest a third of people moved by two or more points on a 0--10 scale after six months. We audit what these repeated answers do to advice. Seven large language models from five providers
gave one-turn allocation advice for 800 real household profiles in which only the risk answer changed. The
recommended stock share moved by 10 to 16 percentage points on average, and by 24 to 35 when the answer changed,
against 0.6 to 4.2 points when the same prompt was repeated; in the laboratory retest the rating alone moved advice
by 11 to 18 points. Without any note, the models respond to the answer about as strongly as
when told that it is very reliable (as strongly for five models, somewhat less for the two Claude
models). If that instructed response is how each model maps attitude to advice, discounting the answer for its
measured reliability, as classical test theory prescribes, would cut the swings by a factor of 2.4 to 3.3. The variation is uneven and largest for clients under
40. Telling the model that the answer is noisy steadies some models, but shifts their weight to the client's
age and pulls cautious clients towards a generic middle allocation: clients unwilling to take any risk
are advised 22\% stocks instead of 4\%. Three extra money questions reduce the swing for some models, with no
statistically detectable change relative to how much advice differs across people. In a scripted test, one
clarifying question removes nearly all of the swing from an answer the client then corrects (33.5
points to 1.9 on average) while keeping the response to a change the client confirms. We argue that same-person retest audits should be a standard
check for AI advisers, and that what to do with an unreliable answer is a question of client autonomy as well as
of statistics.
\end{abstract}


\maketitle

\section{Introduction}

Retail investment advice is moving to software. Robo-advisers already manage portfolios from short online
questionnaires, and general-purpose chat assistants are now asked the same questions directly: how much of my
savings should be in stocks? Regulators require that advice be \emph{suitable}, which in practice means that it
must follow from what the client says about their goals, circumstances and willingness to take risk
\citep{esma,sec-robo,finra,regbi}. Recent audits show that large language models (LLMs) asked for an allocation
lean on one input above all others: the client's stated risk tolerance \citep{ross-lo,fieberg,rumpf}. Rumpf et al.
\citep{rumpf} find that an LLM's advice responds to declared risk tolerance about three times as strongly as
the advice of human advisers, lay and professional.

That reliance would be harmless if the answer were a precise measurement. It is not: answers to risk-tolerance
questions carry large response error \citep{kimball,sahm,mata,schildberg,gillen}, and the same person, asked again,
often answers differently for reasons unrelated to their finances (mood, framing, what they read that morning). An
adviser that maps the answer tightly to the portfolio passes that noise into the advice, so one person on two
different days can receive very different recommendations.

This is a measurement problem in the sense of Jacobs and Wallach \citep{jacobs-wallach}: a construct (the
client's attitude to risk) is operationalised through a noisy instrument (one survey item), and a consequential
decision treats the instrument's output as the construct itself. It is also an accountability problem. A client
cannot see that their allocation would have been 20 points different had they ticked the neighbouring box, and a
deployer that tests its adviser on fixed personas never observes the variation either.

We ask a simple question: \emph{when the same person answers the risk question twice, how much does the AI
adviser's recommendation change?} The test needs no view on what the right advice is: a person whose
circumstances and underlying preferences did not change should not receive very different advice, and if they do,
a deployer, a regulator or a client can reasonably object.

The problem is not specific to AI, and we do not claim that LLMs advise worse than people: any adviser that maps
one noisy answer tightly to a portfolio would show the same swing. The question is whether an adviser treats its
input with the reliability that input actually has. We audit LLMs because they are increasingly
the adviser, and because, unlike a human adviser, the same one can be asked about the same client again with only
the answer changed.

We answer it with public retest data on real people and an audit of seven LLMs from five providers. Our
contributions are:
\begin{enumerate}
\item \textbf{Retest audit.} Prior audits show that LLM advice depends strongly on stated risk tolerance; we give
  the first measurement, on real people's repeated answers, of how much that advice changes when the answer
  changes on retest (Sections~\ref{sec:noise}--\ref{sec:results}), with
  robustness to prompt wording, sample size and a crisis-free six-month retest (Section~\ref{sec:robust}).
\item \textbf{A reliability benchmark} from classical test theory \citep{spearman,kelley}: how much advice
  would move if each model discounted the answer for its reliability. It assumes that the model's response when
  told the answer is very reliable is its response to attitude, and it measures consistency with the model's own
  mapping from attitude to advice, not correct advice (Section~\ref{sec:bench}).
\item \textbf{Distribution of the instability.} The variation is uneven, largest for younger households; the larger swings of the less wealthy shrink once age is held fixed (Section~\ref{sec:who}).
\item \textbf{Three mitigations} and their trade-offs: a noise warning that steadies some models by shifting
  weight to the client's age and raises advice for the most cautious clients, extra questions that add caution
  more than stability, and, in a scripted test, a clarifying question that removes nearly all of the swing from a
  corrected answer while keeping the response to a confirmed change (Section~\ref{sec:fix}), with a discussion of when correcting a stated preference
  overrides the client (Section~\ref{sec:discussion}).
\item \textbf{A reusable audit.} Code and all 77,014 model replies, so that a deployer or regulator can run the
  same-person audit on any adviser (Section~\ref{sec:discussion}), at \url{https://anonymous.4open.science/r/financial-advice-wobble-B279/}.
\end{enumerate}

We keep three claims apart. \emph{Input-induced instability}, advice that changes when only the answer changes, is
measured directly (Sections~\ref{sec:results} and~\ref{sec:robust}). \emph{Amplification of measurement error},
advice that moves more than the answer's reliability justifies, needs a measurement model and an estimate of
how each model maps attitude to advice; Section~\ref{sec:bench} approximates the latter by the model's response
when told that the answer is very reliable, which adds an assumption. \emph{Harm} to clients we do not measure: we observe recommendations, not outcomes.

\section{Related work}\label{sec:related}

\paragraph{LLM and robo financial advice.} Ross and Lo \citep{ross-lo} find that frontier LLMs collapse to
heuristics in investment advice, setting allocations largely from self-reported risk tolerance; Fieberg et al.
\citep{fieberg} report the same pattern across more models. Rumpf et al. \citep{rumpf} compare one LLM's advice
with that of professionals and lay advisers and find it about three times as sensitive to declared risk. Choukhmane et al.
\citep{choukhmane} compare LLM advice with life-cycle benchmarks and measure how much reruns of the same prompt
vary. Other audits examine investment biases \citep{winder}, risk profiling \citep{hens}, personalised advisory
agents \citep{takayanagi} and the risk profiles of the models themselves \citep{cho}. Before LLMs, studies of
robo-advisers showed that they map short questionnaires to portfolios in very different ways
\citep{tertilt}, and that human advisers often give similar portfolios to clients with different profiles
\citep{foerster}; Capponi et al. \citep{capponi} model how a robo-adviser should learn risk preferences through
repeated interaction. None of this work measures how much the client's \emph{own} answer moves on retest, or
what that movement does to the advice. The studies that measure variation (Rumpf et al., Choukhmane et al.) measure
the model's sampling noise, not noise in its input.

\paragraph{Reliability of stated risk tolerance.} Survey measures of risk attitude are validated against
incentivised choices \citep{dohmen}, and reviews and large psychometric studies find them more stable than
behavioural tasks, although single items are only moderately stable over time \citep{mata,schildberg,frey}.
Kimball et al. \citep{kimball} estimate that response error contributes more variance to hypothetical
income-gamble answers than true risk tolerance; Sahm \citep{sahm} finds that persistent differences between people
account for most systematic variation. Correcting for this measurement error changes substantive conclusions
\citep{gillen}. Some change is real: preferences
shifted after the 2008 crisis \citep{guiso,bricker}.

\paragraph{Consistency and auditing of LLMs.} LLM outputs change with prompt features that should not matter
\citep{sclar,mizrahi}, models answer paraphrases inconsistently \citep{elazar}, and even nominally deterministic
settings vary \citep{atil}. Behavioural testing \citep{ribeiro} and psychometric reliability checks
\citep{shu,rottger,dominguez} treat the model as the object being measured. Our design reverses this: the model is
the instrument that consumes a measurement, and the question is how it propagates the measurement's error. The
audit follows the outside-in tradition of algorithmic auditing \citep{metaxa} and is intended for use in internal
audit frameworks \citep{raji}.

\section{How noisy is the answer?}\label{sec:noise}

We use two public retests (Table~\ref{tab:noise}).

\paragraph{SCF 2007--09 panel.} The Survey of Consumer Finances \citep{scf} re-interviewed its 2007 households
in 2009 and asked the same four-level question both times: ``Which of the following statements comes closest to describing
the amount of financial risk that you (and your [husband/wife/partner]) are willing to take when you save or make
investments?'', with answers from
``Take substantial financial risks expecting to earn substantial returns'' to ``Not willing to take any financial
risks''. The question can cover a couple's joint attitude, so we keep the 3,712 households in which the same person
answered both times. Among them, the panel-weighted retest
correlation is 0.47 (95\% interval 0.44 to 0.50). The correlation measures how well the answer keeps people in
order, not how often they repeat it: 41\% chose a different answer and 7.5\% moved two or more of the four levels. The two interviews straddle the 2008 crisis, so part of the change is real \citep{bricker}. But the
average barely moved (1.72 to 1.83 on the 1--4 scale): the change is not a common shift, although a stable average
can also hide real changes in both directions.

\paragraph{Basel--Berlin Risk Study.} The Basel--Berlin study \citep{frey,pedroni} asked the 0--10 financial risk rating
\citep{dohmen} (``How would you rate your willingness to take risks in the following areas? \ldots{} in financial
matters'', from 0, unwilling, to 10, fully prepared) and retested a
subsample six months later, with no crisis in between. In the Berlin retest, the correlation is 0.54; a third of
people moved by two points or more and one in six by three or more. Pooling the Berlin and Basel retests, the 171
people our audit uses, gives 0.57, with 36\% moving two points or more.

\begin{table}[t]
\centering\small
\begin{tabular}{lrll}
\toprule
Sample & $n$ & $r$ & Changed \\
\midrule
SCF panel, 4 levels, 2 years & 3,712 & 0.47 & 41\% \\
Basel--Berlin, 0--10, 6 months (Berlin) & 109 & 0.54 & 34\% ($\geq$2 pts) \\
\quad Basel and Berlin pooled (audit sample) & 171 & 0.57 & 36\% ($\geq$2 pts) \\
\bottomrule
\end{tabular}

\caption{Retest of the stated-risk answer. SCF: panel-weighted, $r$ = 0.47 [0.44, 0.50]. Basel--Berlin: the Berlin
retest subsample, and both retests pooled (the people audited in Sections~\ref{sec:bench} and~\ref{sec:fix}).}
\label{tab:noise}
\end{table}

These values lie within the published range for single risk items over similar gaps \citep{mata,sahm}: in a
laboratory sample and in a representative national sample alike, one stated-risk answer is a noisy input.

\section{Audit design}\label{sec:design}

\paragraph{Profiles.} We drew 800 households from the SCF panel, without replacement and with probability
proportional to the panel weight (200 first, then 600 from the remaining households), among the 2,859 eligible
households where the same person answered both waves and financial assets were at least \$5{,}000. Each
profile gives the household's 2007 age, education, income, net worth and financial assets, the risk question,
and one answer. The two arms of the audit differ \emph{only} in that answer: the person's 2007 answer or their
2009 answer. Holding the facts at their 2007 values isolates the effect of the answer. For the six-month retest we
built profiles for 171 Basel--Berlin retest participants (Berlin and Basel), with country, education, a fixed
age band and savings amount (the public files hold neither age nor wealth), and their first or second rating.
The prompts are in Appendix~\ref{app:prompts}.

\paragraph{Advisers.} Claude Sonnet 5.5, Claude Haiku 5.5, GPT-6.1 Sol, GPT-6 Luna, Gemini 3.8 Flash, Grok 4.7 and
Mistral Large 4, queried through one API router with default sampling settings, one profile per request (exact
identifiers, dates and settings in Appendix~\ref{app:prompts}). Each
model returns the recommended percentage of financial assets in stocks as JSON. The first 200 SCF households were
asked twice per prompt, which gives a same-prompt control; the other 600 once.

\paragraph{Measures.} The \emph{swing} is the mean absolute difference in recommended stock share between a
person's two answers. The \emph{spread} is the standard deviation of advice across people; their ratio is the swing
per unit of personalisation. \emph{Points per level} is the slope of the change in advice on the change in answer,
which we compare with the same slope in real holdings (a panel-weighted regression of the equity share of
financial assets on the answer and the same facts, over all 2,859 panel households that meet the same
sample conditions).

\paragraph{Inference.} Because the profiles were drawn with probability proportional to the weight, plain means
over profiles approximate weighted means over the eligible households; we report plain means. All SCF values come
from the first of the survey's five imputations (implicates). Intervals are percentile intervals from 2,000
bootstrap draws over the audited households (or people), using the same draws for every model and condition so
that differences are paired. They are conditional on the first implicate and on the eligible panel (households with at least \$5{,}000 in
financial assets), and omit imputation and survey-design variance \citep{scf}.

\section{Same person, different advice}\label{sec:results}

\begin{figure}[t]
\centering
\includegraphics{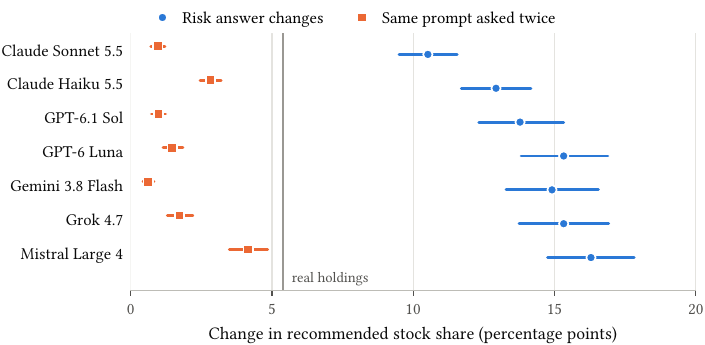}
\Description{Dot plot: for each of the seven models, the mean change in recommended stock share when only the risk answer changes (10 to 16 points) and when the same prompt is asked twice (under 5 points), with a vertical line at the change implied by real holdings.}
\caption{Mean change in recommended stock share when only the risk answer changes, against the same prompt asked
twice. Grey line: the change implied by real holdings. 95\% bootstrap intervals.}
\label{fig:swing}
\end{figure}

Changing only the risk answer moves the recommended stock share by 10.5 to 16.3 percentage points on average,
depending on the model, and by 24 to 35 points for the households whose answer changed (Table~\ref{tab:wobble},
Figure~\ref{fig:swing}). Between 23\% and 35\% of households see their advice move by 20 points or more. Asking the
same prompt twice moves the advice by only 0.6 to 4.2 points, so the swing comes from the input, not from sampling.

\begin{table}[t]
\centering\small
\resizebox{\linewidth}{!}{\begin{tabular}{lllll}
\toprule
 & Swing & Swing if answer changed & Same prompt twice & pp per level \\
\midrule
Claude Sonnet 5.5 & 10.5 {\scriptsize [9.5, 11.6]} & 24.1 {\scriptsize [22.5, 25.5]} & 1.0 {\scriptsize [0.7, 1.2]} & 19.5 {\scriptsize [18.1, 20.9]} \\
Claude Haiku 5.5 & 12.9 {\scriptsize [11.7, 14.2]} & 27.4 {\scriptsize [25.2, 29.4]} & 2.8 {\scriptsize [2.5, 3.2]} & 22.5 {\scriptsize [20.8, 24.3]} \\
GPT-6.1 Sol & 13.8 {\scriptsize [12.3, 15.3]} & 32.2 {\scriptsize [29.9, 34.5]} & 1.0 {\scriptsize [0.7, 1.2]} & 26.7 {\scriptsize [24.8, 28.7]} \\
GPT-6 Luna & 15.3 {\scriptsize [13.8, 16.9]} & 34.7 {\scriptsize [32.1, 37.2]} & 1.5 {\scriptsize [1.1, 1.9]} & 28.4 {\scriptsize [26.2, 30.6]} \\
Gemini 3.8 Flash & 14.9 {\scriptsize [13.3, 16.6]} & 35.1 {\scriptsize [32.2, 37.7]} & 0.6 {\scriptsize [0.4, 0.8]} & 28.5 {\scriptsize [26.2, 30.8]} \\
Grok 4.7 & 15.3 {\scriptsize [13.7, 16.9]} & 34.9 {\scriptsize [32.4, 37.4]} & 1.7 {\scriptsize [1.3, 2.2]} & 28.0 {\scriptsize [25.8, 30.3]} \\
Mistral Large 4 & 16.3 {\scriptsize [14.8, 17.8]} & 34.3 {\scriptsize [31.7, 36.7]} & 4.2 {\scriptsize [3.5, 4.8]} & 27.8 {\scriptsize [25.6, 30.0]} \\
\midrule
Real holdings & 5.4 {\scriptsize [4.4, 6.4]} & & & 10.4 {\scriptsize [8.5, 12.2]} \\
\bottomrule
\end{tabular}
}
\caption{Change in recommended stock share when only the risk answer changes (800 SCF panel households; same
prompt twice: the first 200). Percentage points; 95\% bootstrap intervals. Real holdings: panel-weighted
regression over all 2,859 panel households, same facts.}
\label{tab:wobble}
\end{table}

Per answer level, the models move advice by 19.5 to 28.5 points; how much of this the answer's reliability can
justify is the subject of Section~\ref{sec:bench}. As a rough external reference only, households whose answer
differs by one level hold, other things equal, 10.4 points [8.5, 12.2] more or less in equities, so the models
react 1.9 to 2.7 times as strongly as real holdings. The comparison is weak (households are inert, and some of the
gap may be good advice), and no conclusion rests on it.

The ordering of models is stable across analyses: Claude Sonnet 5.5 swings least, Mistral Large 4 most. The
spread between them is small next to the gap between any model and the reliability benchmark
(Section~\ref{sec:bench}).

\section{Robustness}\label{sec:robust}

\paragraph{Outside the crisis.} The SCF panel straddles 2008. In the Basel--Berlin retest, six months apart with no
crisis, the rating alone moves advice by 11.2 to 17.7 points, as much as the two-year SCF swing, so the result does not
depend on the crisis, although neither retest fully separates response error from genuine preference change.

\paragraph{Not the wording.} On the first 100 households we re-asked every model with three rewritten prompts:
a fiduciary system prompt (``you must act in the client's best interest''), the client describing themselves in
the first person, and a stocks/bonds/cash split instead of one number. The swing stays at 11.8 to 20.5 points (Table~\ref{tab:robust}).
Paired with the original prompt on the same households, every rewording changes a model's swing by between
$-1.4$ and $+2.5$ points; five of the 21 paired differences exclude zero, the largest Mistral Large 4 in the client's voice ($+2.5$
[0.5, 4.7]), small next to the swing itself.

\paragraph{Not the sample.} The first 200 households alone give 11.2 to 18.1 points; the 600 added later in the
same way give 10.3 to 15.7. The first sample sat at the high end, but the ranking of models and the gap to real
holdings do not depend on which set is used.

\begin{table}[t]
\centering\small
\resizebox{\linewidth}{!}{\begin{tabular}{llllll}
\toprule
 & \multicolumn{4}{c}{Prompt wording (100 households)} & 800 households \\
\cmidrule(lr){2-5}\cmidrule(lr){6-6}
 & Original & Fiduciary & Client voice & Three assets & Original \\
\midrule
Claude Sonnet 5.5 & 11.8 {\scriptsize [8.8, 14.9]} & 13.4 {\scriptsize [10.1, 17.0]} & 11.8 {\scriptsize [8.7, 15.2]} & 13.2 {\scriptsize [9.8, 16.7]} & 10.5 {\scriptsize [9.5, 11.6]} \\
Claude Haiku 5.5 & 14.3 {\scriptsize [10.7, 18.1]} & 14.6 {\scriptsize [10.6, 18.8]} & 12.9 {\scriptsize [10.1, 16.1]} & 14.6 {\scriptsize [10.9, 18.4]} & 12.9 {\scriptsize [11.7, 14.2]} \\
GPT-6.1 Sol & 16.6 {\scriptsize [12.1, 21.0]} & 15.6 {\scriptsize [11.4, 20.3]} & 16.4 {\scriptsize [11.9, 21.1]} & 16.2 {\scriptsize [12.1, 20.8]} & 13.8 {\scriptsize [12.3, 15.3]} \\
GPT-6 Luna & 17.4 {\scriptsize [12.7, 22.1]} & 16.8 {\scriptsize [12.3, 21.7]} & 16.9 {\scriptsize [12.3, 21.9]} & 18.0 {\scriptsize [13.6, 22.6]} & 15.3 {\scriptsize [13.8, 16.9]} \\
Gemini 3.8 Flash & 17.2 {\scriptsize [12.4, 22.4]} & 17.6 {\scriptsize [12.7, 22.9]} & 18.1 {\scriptsize [13.0, 23.4]} & 16.9 {\scriptsize [12.2, 21.7]} & 14.9 {\scriptsize [13.3, 16.6]} \\
Grok 4.7 & 17.1 {\scriptsize [12.4, 22.1]} & 16.8 {\scriptsize [12.0, 21.5]} & 17.6 {\scriptsize [13.0, 22.7]} & 17.6 {\scriptsize [13.3, 22.3]} & 15.3 {\scriptsize [13.7, 16.9]} \\
Mistral Large 4 & 18.2 {\scriptsize [13.5, 23.0]} & 19.2 {\scriptsize [14.6, 24.0]} & 20.5 {\scriptsize [15.7, 25.4]} & 19.6 {\scriptsize [15.2, 24.3]} & 16.3 {\scriptsize [14.8, 17.8]} \\
\bottomrule
\end{tabular}
}
\caption{Swing (percentage points of recommended stock share) when only the risk answer changes, under three
rewordings of the prompt (first 100 households, one answer per prompt; Original is the first repeat of the main
run) and on all 800 households. 95\% bootstrap intervals.}
\label{tab:robust}
\end{table}

\begin{figure}[t]
\centering
\includegraphics{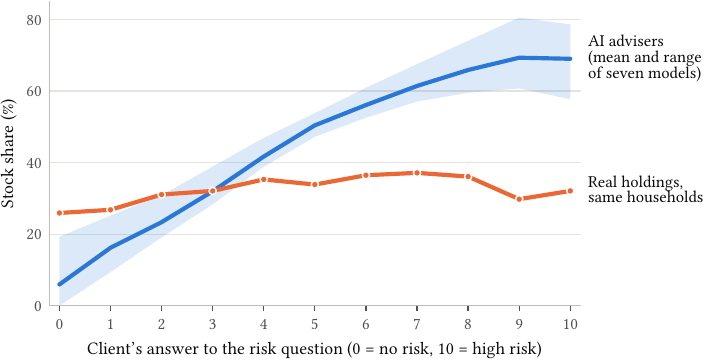}
\Description{Line chart: the models' recommended stock share (mean line and range band) rises steeply across the 0 to 10 risk answer, while the share implied by real holdings for the same households stays between 26 and 37 percent.}
\caption{Recommended stock share at each answer to the 0--10 risk rating for 100 SCF 2022 households whose other
facts are held fixed: mean of the seven models (blue line) and the range across models (band). Orange: the equity
share that real SCF 2022 holdings imply for the same households at each answer (survey-weighted regression on the
answer and the same facts).}
\label{fig:curve}
\end{figure}

\paragraph{The full response curve.} To see the mapping directly, we described 100 households from the SCF 2022 \citep{scf2022}
cross-section eleven times each, once at every answer to the 0--10 rating, keeping their other facts fixed
(Figure~\ref{fig:curve}). Advice rises by 5.3 (Claude Haiku 5.5) to 7.8 (Mistral Large 4) points per answer
point, almost linearly through the middle of the scale before flattening at the top, and advice at an answer of 10 sits 48 to 72 points above advice at 0. At an answer of 0, GPT-6.1
Sol and Gemini 3.8 Flash put all 100 clients in 0\% stocks, whatever their age, wealth or horizon. In real SCF 2022
holdings, with the same facts controlled, one point on the scale goes with 0.9 points of equity share: for these
100 households, real holdings imply a stock share between 26\% and 37\% across the whole scale.

\paragraph{Sampling and parsing.} We sent 77{,}014 requests in all (Table~\ref{tab:accounting}), 29{,}750 of them for the
clarification test of Section~\ref{sec:fix}. Four replies
could not be read, three in the main audit and one with the noise note; 341 attempts that hit a rate limit or a
provider error, or repeated a request, were retried. A first Basel--Berlin run of 7{,}182 requests showed the rating
one point high (the study's scores file stores the 0--10 rating as 1--11); it is released but not used.

\paragraph{Not sampling randomness.} The main runs use each provider's default temperature. Re-asking the first 100
households at temperature 0 produces no statistically detectable change in the swing: 12.1 to 17.6 points, against 11.8 to 18.2 at the default,
and no model's paired difference is distinguishable from zero ($-0.6$ to $+0.5$ points). Some providers may not
apply the setting to reasoning models, but the result matches the same-prompt control: the swing comes from the
answer, not from sampling.

\section{How much should advice move?}\label{sec:bench}

Some movement is legitimate: if attitude matters for advice and the answer carries information about it, advice
should respond. The question is how much. Classical test theory gives an answer that keeps each model's own view of
how advice should depend on attitude.

Let the stated answer be $x = t + e$, with true attitude $t$ and noise $e$ independent of $t$ and of the stable facts
$Z$ in the profile. Let $\rho = \mathrm{var}(t)/\mathrm{var}(x)$ be the answer's reliability (the retest
correlation) and $R^2 = \mathrm{var}(\mathrm{E}[x \mid Z])/\mathrm{var}(x)$ the share of the answer predicted by the
facts. The best linear estimate of the attitude given the answer and the facts is Kelley's true-score estimate
\citep{kelley}, conditioned on $Z$:
\begin{equation}
\hat t = m(Z) + \lambda\,\bigl(x - m(Z)\bigr), \qquad \lambda = \frac{\rho - R^2}{1 - R^2},
\end{equation}
where $m(Z)$ is the attitude predicted from the facts (derivation in Appendix~\ref{app:lambda}). The \emph{benchmark adviser} is a linear
advice rule of the model (stock share on the answer and the facts) fed $\hat t$ instead of $x$. The
\emph{original benchmark} fits the rule to the model's plain-prompt answers, treating its response to the observed
answer as its response to attitude; for a model that already discounted the answer, it would discount the answer
a second time. The \emph{corrected benchmark} fits the rule to the model's answers when told that the answer is very
reliable (below).

In the SCF panel, $\rho = 0.47$ and the facts explain $R^2 = 0.21$ of the answer, so $\lambda = 0.33$ [0.28, 0.38]:
a one-level change in the answer should move the estimate of attitude by a third of a level. On the 800
households, the original benchmark swing is 3.3 to 4.8 points; the observed swing is 3.0 to 3.4 times larger.
Measured against personalisation, the original benchmark's swing/spread ratio on the 800 households is 0.17 to
0.24, against 0.43 to 0.52 for the plain prompt (Table~\ref{tab:ratio} in Appendix~\ref{app:lambda} shows the first 200, where both notes were
tested: 0.19 to 0.27 for the original benchmark and 0.24 to 0.28 for the corrected one, against 0.46 to 0.56).

Because the original benchmark's rule is fitted to the model's own plain answers, the ratio of observed to
benchmark swing is close to $1/\lambda$ by construction; for the corrected benchmark it is close to $k/\lambda$,
where $k$ is the share of its instructed response to the answer that the model keeps without the note
(Appendix~\ref{app:lambda}). Neither is an external standard of correct advice: the swings of
Sections~\ref{sec:results} and~\ref{sec:robust} do not depend on it, while the claim that they are excessive does,
and holds only under five assumptions: (i) the answer is attitude plus noise, with the noise independent of the
attitude; (ii) the noise is also independent of the facts; (iii) the model's advice is linear in the answer and the
facts; (iv) the retest correlation measures the answer's reliability; and (v) the model's response when told that the
answer is very reliable is its undiscounted response to attitude (for the original benchmark: its response to the
plain answer is), which we probe next but cannot verify.

\paragraph{The corrected benchmark.} We asked every model about the first 200 households again, with one line
where the noise note of Section~\ref{sec:fix} goes: ``answers to this risk question are very reliable; when the
same people answer it again a couple of years later, almost all choose the same statement'' (2{,}800 requests).
Plain-prompt sensitivity remains close to sensitivity under this explicit high-reliability instruction. Five
models keep 0.96 to 1.00 of their instructed answer slope without the note (GPT-6.1 Sol, GPT-6 Luna, Gemini 3.8
Flash, Grok 4.7, Mistral Large 4); the two Claude models keep 0.89 [0.86, 0.93] (Claude Haiku 5.5) and 0.77
[0.74, 0.79] (Claude Sonnet 5.5), where classical test theory would keep 0.33. The note leaves mean advice within 3
points. This does not show that the instructed response is the model's undiscounted response to attitude. A model
could ignore the note and keep discounting internally, in which case both benchmarks discount twice and overstate
the excess, or read the note as a request to lean harder on the answer, in which case the corrected benchmark
understates it. That the same models do respond to the opposite note (Section~\ref{sec:fix}) shows that they read
such notes, not how they map attitude to advice; reading the instructed response as that mapping is assumption
(v). Under it, the observed swing is 2.4 [2.1, 3.0] (Claude Sonnet 5.5) to 3.3 [2.8, 4.2] (Mistral Large 4) times
the corrected benchmark, against 3.1 to 3.4 times the original benchmark on the same households; the paired
difference is $-0.74$ [$-0.94$, $-0.61$] for Claude Sonnet 5.5, $-0.37$ [$-0.55$, $-0.23$] for Claude Haiku 5.5 and
between $-0.14$ and $+0.01$ for the others (Table~\ref{tab:byrho}).

\begin{figure}[t]
\centering
\includegraphics{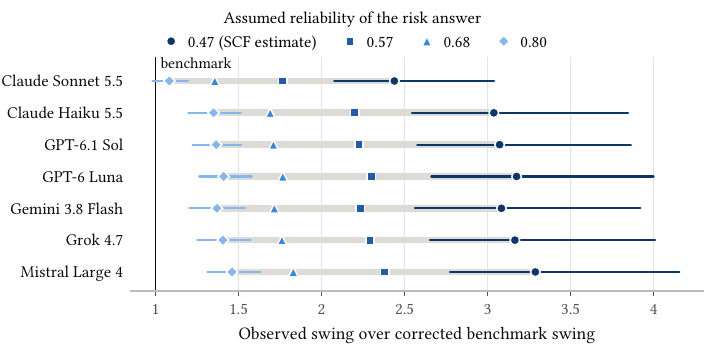}
\Description{Dumbbell plot: for each model, a grey bar joins the observed swing divided by the corrected benchmark swing at four assumed reliabilities; the ratio falls from about 2.4 to 3.3 at 0.47 to about 1.1 to 1.5 at 0.80, with a vertical line at 1 for the benchmark.}
\caption{Observed swing over the corrected benchmark swing for each model, at the estimated reliability of the
answer (0.47) and at three higher assumed reliabilities, on the first 200 SCF households ($R^2$ and $m(Z)$ held at
their estimates). 1 is the benchmark. Grey bar: the span from the most generous reliability (0.80, left) to the
estimate (0.47, right). Thin lines: 95\% bootstrap intervals at those two ends; every interval, both benchmarks and a
reliability of 0.6 in Table~\ref{tab:byrho}.}
\label{fig:excess}
\end{figure}

How reliable is the answer? Classical test theory treats every difference between the two answers as noise, but
risk preferences can also move with circumstances (background risk, income shocks, a need for cash), and advice
should follow such shifts. If some 2007--09 change was real, the SCF $\rho$ understates the answer's reliability. The Basel--Berlin
retest, six months apart with no crisis, gives $\rho = 0.57$, and stated risk-propensity measures retest at 0.68 on
average in the same study \citep{frey}. If the answer's short-term reliability lies between these values, then 60\%
to 80\% of the variance of the 2007--09 answer change is noise, and real updating accounts for the rest. Figure~\ref{fig:excess} and Table~\ref{tab:byrho} recompute both benchmarks at these and higher reliabilities on the same
households, holding $R^2$ and $m(Z)$ at their estimates. Against the corrected benchmark the models swing 1.8 to
2.4 times as much at $\rho = 0.57$, 1.6 to 2.2 times at a generous 0.6, and 1.4 to 1.8 times at 0.68. At
$\rho = 0.8$, above any single-item retest we know of, the ratio is 1.3 to 1.5 for six models and 1.08 [0.98, 1.20]
for Claude Sonnet 5.5, whose swing would then be statistically compatible with the benchmark. How excessive the swings are thus
depends on the answer's reliability as well as on assumption (v). The four answers are ordered categories, not equally spaced numbers.
Redoing the whole benchmark on a latent continuous scale, with the answers as coarse cuts of it (a polychoric
retest correlation, an ordered probit for the facts, and each model's rule refitted on the latent score;
Appendix~\ref{app:lambda}), gives a latent reliability of 0.56 [0.52, 0.60], $R^2 = 0.26$ and
$\lambda = 0.40$ [0.34, 0.46]; the models then swing 2.5 to 2.8 times the original benchmark, and 2.0 [1.7, 2.4]
(Claude Sonnet 5.5) to 2.7 [2.3, 3.4] (Mistral Large 4) times the corrected benchmark on the first 200 households.

In the Basel--Berlin profiles, which hold no other informative facts ($R^2 \approx 0$), $\lambda = \rho = 0.57$.
Without facts to lean on, the best any adviser can do is shrink all advice towards the middle; the benchmark
swing/spread ratio (0.57 to 0.61) is then close to the plain prompt's (0.59 to 0.65), because both swing and spread
shrink together. The benchmark therefore suggests that telling the model the answer is noisy should help more when
the profile holds other stable facts. Section~\ref{sec:fix} is consistent with this, but the two samples differ in
more than the facts, so the comparison is not a controlled test.

\section{Who bears the wobble?}\label{sec:who}

\begin{figure}[t]
\centering
\includegraphics{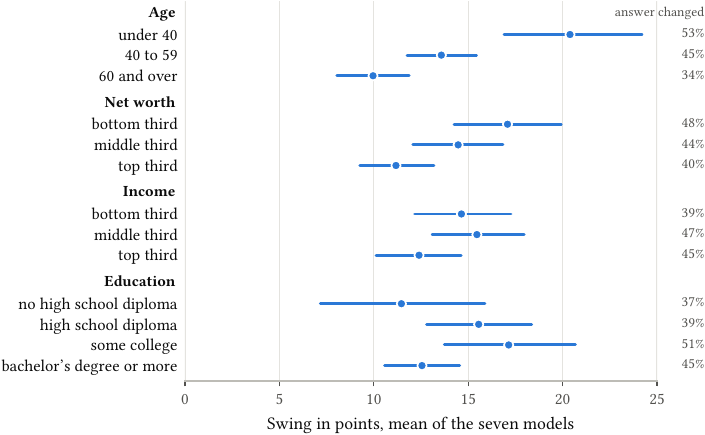}
\Description{Dot plot: swing by household group, largest for clients under 40 and for the less wealthy, with the share of each group whose answer changed listed on the right.}
\caption{Swing by household group (800 SCF panel households, 2007 characteristics), averaged over the seven
models, with 95\% bootstrap intervals. Right: the share of the group whose answer changed, among all 2{,}859
eligible panel households (panel-weighted). Details in Table~\ref{tab:who}.}
\label{fig:who}
\end{figure}

Noise in advice is not only inefficient; it can be unequal. What we compare across groups is the instability of
advice, not its level: recommending more stocks to younger clients can be justified by their horizon, but if a young
client's underlying preference has not changed, advice that moves further for them has no such justification. Two
things determine how much a group is affected: how often its members' answers change (exposure) and how far the
advice moves when they do (sensitivity). Figure~\ref{fig:who} breaks the 800 households down by their 2007 net
worth, income, age and education (Table~\ref{tab:who} in Appendix~\ref{app:who} adds the audited households' shares, the swing if the answer changed and the range across models). All of them held at least \$5,000 in financial assets in 2007
(Section~\ref{sec:design}), so the least wealthy households, who may be the most exposed, are not covered.

The gradient is steepest by wealth and age. Households in the bottom third of net worth see an average swing of
17.1 points [14.2, 19.9], against 11.2 [9.3, 13.2] for the top third; households under 40 see 20.4 [16.9, 24.2],
against 10.0 [8.0, 11.9] for those 60 and over. Both exposure and sensitivity contribute. The less wealthy and the
young change their answer more often: 48\% and 53\% of all eligible households, against 40\% and 34\% (among the
audited households, 43\% and 50\%, against 36\% and 31\%). And when the answer does change, the
models move their advice further: among households whose answer moved, a one-level change moves advice by about
32 points for the bottom wealth third and for the under-40s, against 24 to 26 points for the top third and the
over-60s (the swing if changed in Table~\ref{tab:who} divided by the average number of levels moved), although the size of the answer change is the same across groups (1.1 to 1.2 levels). The response curves of
Section~\ref{sec:robust} test the sensitivity part directly, because there every household is asked at every
answer with its facts fixed. The age gradient is built into the models' rules: advice moves 8.4 points per point of
the 0--10 scale for clients under 40 and 5.2 for those 60 and over, because the models cap older clients' stock
share (at an answer of 10, 85\% against 55\%). The curves show no wealth gradient (6.4 to 6.9 points per point
across wealth thirds). The wealth gradient in the panel therefore reflects exposure, age (the bottom wealth third
is younger, median age 40 against 55 for the top third), and where answers move: 62\% of the bottom third's answer
changes involve the most cautious level, whose advice sits near 0\%, against 39\% for the top third.
A regression of each household's swing (averaged over the models) on age, net worth, income and education
together separates the two. Alone, each tenfold increase in net worth goes with a swing 2.2 points smaller
[0.9, 3.7]; next to the other facts, 0.8 points [$-0.6$, 2.4], indistinguishable from zero, while age keeps its
effect: 2.5 points less per decade [1.6, 3.5]. Income and education add nothing distinguishable from zero. Adding
how far the answer moved and whether the move involves the most cautious level leaves age at 1.7 points per decade
[1.4, 2.1]; wealth then shows no effect distinguishable from zero, and income, if anything, slightly raises the
swing (0.7 points per tenfold [0.2, 1.4]).

Less wealthy households do receive less stable advice, but the gap shrinks substantially once age, income and
education are held fixed, and the adjusted association with net worth is indistinguishable from zero, so we do
not read the unadjusted wealth gradient as an independent effect of wealth; the robust association is with age.
Education shows no monotone pattern; by income, only the top third stands out, with smaller swings. Younger and less wealthy
clients are arguably those for whom advice matters most: younger investors have the longest horizon over which an
allocation compounds, and less wealthy households are the target market of low-cost digital advice. Whether the
disparity harms them depends on outcomes we do not observe; what the audit shows is that the instability is not
spread evenly.

\section{Can the advice be steadied?}\label{sec:fix}

We tested two cheap mitigations that a deployer could add to a prompt today (Table~\ref{tab:fixes} in Appendix~\ref{app:fix}), and, in a
scripted test, a third that involves the client: asking.

\paragraph{A note that the answer is noisy.} One added line (``answers to this risk question are not very
reliable; when the same people answer it again a couple of years later, about 4 in 10 choose a different
statement'') cuts the SCF swing by 4.0 to 8.7 points for six models; GPT-6.1 Sol moves by 1.4. The swing/spread
ratio falls clearly for Claude Sonnet 5.5 ($-0.18$ [$-0.24$, $-0.13$]), Mistral Large 4 ($-0.17$), Claude Haiku 5.5
($-0.11$) and Gemini 3.8 Flash ($-0.07$), which means these models shift weight from the answer to the stable facts
rather than merely flattening advice. Fitting each model's advice rule (Section~\ref{sec:bench}) with and without
the note shows where the weight goes. The note cuts the slope on the answer to 0.37 to 0.46 of its plain value for
Claude Sonnet 5.5, Claude Haiku 5.5, Grok 4.7 and Mistral Large 4, to 0.64 to 0.71 for GPT-6 Luna and Gemini 3.8
Flash, and to 0.91 for GPT-6.1 Sol. The weight moves mostly to age: for Mistral Large 4 the spread across people
of the part of advice due to age doubles (standard deviation 8 to 16 points), while the part due to financial
assets barely moves (3.0 to 3.3). Claude Sonnet 5.5 with the note swings 5.2 points, close to its corrected benchmark swing of 4.6 [3.5, 5.7] points.
But the note has a side effect the benchmark does not: it raises average advice by 3.1 to 8.8 points for six
models, while the note that the answer is very reliable moves it by less than 3 (Section~\ref{sec:bench}). The mechanism shows up by answer level. With the note, clients who said they were not willing to take any
financial risk are advised 22\% stocks on average instead of 4\%, while clients at the two riskiest levels lose 6
to 8 points. The models pull every client towards a generic middle allocation, not towards what the client's own
facts predict (which is what the benchmark does, moving average advice by only about 2 points). Because cautious
answers are common and their advice sits near 0\%, the pull raises the average.

The note behaves differently in the Basel--Berlin profiles, which hold little besides the rating to lean on, as
the benchmark suggests (Section~\ref{sec:bench}); the two samples differ in more than the facts, so this is not a
controlled test. There the note shrinks the swing for most models but the spread across people as well: the ratio falls
for no model and rises for Claude Haiku 5.5 ($+0.15$ [0.06, 0.25]) and Claude Sonnet 5.5 ($+0.07$ [0.01, 0.12]).
Claude Haiku 5.5 lands near its benchmark swing (7.9 against 7.4 points), but the note raises its average
advice by 16.1 points [14.4, 17.6]; for six of the seven models it raises average advice by 2.1 to 16.1 points.
The same pull is at work: more than half of these ratings are 0 to 2 on the 0--10 scale, and for them Claude
Haiku 5.5's advice rises by 25 points with the note.

\paragraph{Three money questions.} Adding the client's answers to three everyday investing questions from the same
study (how likely they would be to invest 10\% of income in a growth fund, 5\% in a conservative stock, 10\% in
government bonds) cuts the raw swing by 1.5 to 5.4 points for six models (Gemini 3.8 Flash's interval includes zero), but it narrows
the spread across people about as much, so the swing/spread ratio shows no statistically detectable change for six models (their
intervals include zero, which does not establish that the ratio is the same; Mistral Large 4's falls by 0.07
[0.00, 0.15]). It also lowers average advice by 5.8 to 12.1 points.
More questions of this kind make advice more cautious and less responsive overall, but not detectably more
stable relative to how much people differ. These items ask about specific products and are not parallel
measurements of the same attitude, so the result does not show that a validated multi-item scale would fail; that
remedy (Section~\ref{sec:discussion}) is untested here.

\paragraph{A clarifying question.} The note and the money questions apply to every client alike, so they
cannot steady advice for a client whose answer was noise without also damping it for a client whose attitude
really changed. A human adviser would ask. We tested asking in a scripted setting, with the plan, prompts and
analysis code committed before any answer was collected (Appendix~\ref{app:fix}). Each of the 200 SCF
households keeps its real facts, with its 2007 answer shown as the previous one. In scenario A, today's answer
repeats it and the client confirms that nothing has changed; in B, today's answer is one level away and the
client, when asked, says they picked the wrong statement; in C, the answer is as in B and the client says the
change is deliberate. The models advised at once, with the noise note, after an exchange in which the adviser asks
one fixed question and the scripted reply follows, and with that question and reply in a one-turn prompt
(28{,}000 requests, \$59.88). We measure the \emph{unwanted swing} $S = |a_B - a_A|$ and the \emph{response to a
confirmed change} $R$, the move from $a_A$ to $a_C$ in the direction of the change. The tests, fixed in advance, were that
the exchange cut $S$, kept at least 0.8 of the direct $R$ (lower 95\% bound), and moved advice in A by less than
3 points (90\% interval).

\begin{figure}[t]
\centering
\includegraphics{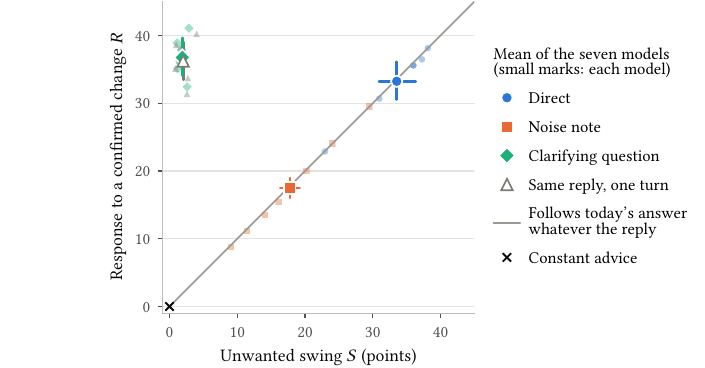}
\Description{Scatter plot of unwanted swing against response to a confirmed change: direct advice sits on the diagonal near 33 points, the noise note about halfway down it, and the clarifying question and its one-turn version at the top left, near 2 points of swing and 37 of response.}
\caption{Clarification test, 200 SCF households. $S$: how far advice moves when today's answer differs from the
previous one and the client then says the previous answer still holds. $R$: how far it moves in the direction of a
change the client confirms. Large marks: mean of the seven models, 95\% bootstrap intervals; small marks: each
model (Table~\ref{tab:clarify}). Top left is best.}
\label{fig:clarify}
\end{figure}

Asked at once (Figure~\ref{fig:clarify}), the models follow today's answer; B and C are then the same prompt,
so the swing equals the response: a one-level change, made for every household here, moves advice by 33.5 points on average (22.9 to 38.1 by
model). The note moves the models down the diagonal. It cuts the swing to 17.8 points but cuts the response to a
confirmed change as much, keeping 0.52 [0.50, 0.54] of it (0.36 to 0.88 by model). The exchange moves every model
to the top left: the swing falls to 1.9 points [1.7, 2.1] (1.1 to 2.9 by model), and the response grows to 1.12
[1.11, 1.14] times its direct value (1.02 to 1.42, the most for Claude Sonnet 5.5, whose direct response is the
smallest). Every model passes the swing and responsiveness tests. Six pass the test on average advice; Claude
Sonnet 5.5 advises the stable client 2.8 points less in stocks (90\% interval $-3.5$ to $-2.1$;
Table~\ref{tab:clarify}). The same question and reply in a one-turn prompt do almost as well ($S$ 2.0, $R$ 36.2),
so the information, not the conversation, steadies the advice. This is a best case: the replies are scripted and
unambiguous, the adviser is handed the question, and the test does not say which real answer changes were noise.

The lesson is that mitigation by prompt is model-specific and has side effects. A deployer cannot assume that
telling a model about measurement error makes it use the answer correctly; it has to measure. Information from
the client works where the warning does not: in the scripted test it separates a mistaken answer from a real
change. Note that the side effect of the note is itself a judgement about the client: an adviser that moves a client who declined all
risk from 4\% to 22\% stocks has overridden what the client said (Section~\ref{sec:discussion}).

\section{Discussion}\label{sec:discussion}

\paragraph{For deployers: audit the same person twice.} Persona-based testing, which varies the profile and
checks that advice moves sensibly, cannot reveal this problem; it rewards exactly the sensitivity that amplifies
noise. The same-person audit is cheap: pair each test profile with its retest answer (from a panel such as the
SCF, or from the deployer's own returning clients), report the swing and the swing/spread ratio as the primary measures,
and the benchmark as an assumption-dependent reference: a swing far above it is a warning sign that calls for a
closer look at how the adviser uses the answer. The audit and its checks took 47{,}264 requests and about \$98 in API fees, and the clarification test
29{,}750 requests and \$64 (Table~\ref{tab:accounting}).

\paragraph{For design: do not let one answer drive the allocation.} Advice should rest on stable facts and on
several measurements of attitude, combined with weights that reflect their reliability. Under the corrected benchmark of
Section~\ref{sec:bench} and its assumptions, advice on the SCF profiles would respond to a one-level change in the answer between a
third and two-fifths as much as current models do; that is a standard of consistency, not of correct advice. Three remedies follow; only the
second is tested here, and only with scripted replies. First, measure attitude with several items and average them: by the Spearman--Brown
formula, three parallel items with the single item's reliability of 0.47 would reach 0.73. Second, when the answer
differs from the client's earlier answer or conflicts with their facts, ask a follow-up question rather than
silently overriding it; in our scripted test, one question with a clear reply steadied every model
(Section~\ref{sec:fix}).
A deployer that does discount the answer, for example by passing a shrunken score such as the benchmark's
$\hat t$, should do so in the pipeline, where it can be checked, rather than in the prompt, which
Section~\ref{sec:fix} shows also shifts mean advice, and should be able to explain it to the client. Third, for
returning clients, keep the history of answers and use their average.

\paragraph{For clients: steadier without overriding.} Steadier advice is not automatically better advice. A risk
question measures willingness to take risk, which is the client's to state; age, income and wealth bear on the
capacity to take it. Pulling the answer towards what the facts predict treats the stated willingness as a noisy
reading of something the adviser can estimate better. The noise note shows what that looks like: clients who said
they would take no financial risk were advised 22\% stocks instead of 4\%. For a client whose caution is real,
that is an override of a stated preference; for one whose answer was noise, it is a correction; and one answer
cannot tell the two apart. In the clarification test the note kept half of the response to a change the client
confirmed; one clarifying question kept all of it. The remedies that respect
both cases gather more information from the client (more
items, a follow-up question, the client's history), so that the adviser is steadier because it knows more about
the client, not because it trusts the client less.

\paragraph{For regulation: suitability needs reliability.} Suitability rules require advisers to collect reliable
information and to check its consistency \citep{esma}, and robo-adviser guidance stresses the design of the
questionnaire \citep{sec-robo,finra}. These rules were written for questionnaires feeding transparent mappings. If deployed advisers behave like the
one-turn advisers audited here, the reliability requirement needs an operational check: an assessment of
suitability should report how much the recommendation would change if the client answered again. This is an
implication to test, not a finding, and we do not claim that current rules already require it. Because the instability is largest for young clients, and the less
wealthy see larger swings before age is taken into account, it also bears on the populations that digital advice
is meant to serve.

\paragraph{What we do not claim.} We do not claim to know the right allocation for any household, and we do not
claim the models are worse than human advisers, whose recommendations also respond to stated risk tolerance \citep{rumpf,foerster} and are
often only weakly tailored to the client \citep{foerster}. We do not measure harm: a 10 to 16 point swing is large, but we observe recommendations, not suitability
violations or investment outcomes. We do not claim that steadier advice is better for every client (above). The central
claim is narrower and measured directly: in one-turn allocation advice, the same client's repeated answers move the
recommendation by 10 to 16 points on average, and the instability falls unevenly. Whether that is more than the
answer's reliability justifies is a secondary claim, conditional on the benchmark's assumptions: if each model's
response when told that the answer is very reliable is its response to attitude, the models move advice 2.4 to 3.3
times as much as discounting for the measured reliability would, and at a reliability of 0.8 six models still
swing 1.3 to 1.5 times as much, while Claude Sonnet 5.5 (1.08 [0.98, 1.20]) is statistically compatible with the
benchmark.

\section{Limitations}\label{sec:limits}

There is no ground truth for the right advice. The benchmark rests on the assumptions of Section~\ref{sec:bench},
in particular that each model's response under the very-reliable note is its response to attitude; the excess is
understated if the note makes a model lean harder on the answer, and overstated if the model keeps discounting
internally. The excess also shrinks as the assumed reliability rises: at
$\rho = 0.8$, six models remain above the benchmark (1.3 to 1.5) while Claude Sonnet 5.5 (1.08 [0.98, 1.20]) is
statistically compatible with it. It measures consistency with each model's own
rule, not correctness. The SCF panel straddles the 2008 crisis, so some answer changes are real; the
Basel--Berlin retest addresses this but its profiles lack age and wealth (the public files do not hold them), so
they rest on the rating alone (on its 0--10 scale in every result reported here). Because the two samples also differ in population,
country, scale and wording, the contrast between the mitigations' effects in them is suggestive, not a controlled
comparison. Our intervals reflect the sampling of households and the models' answers, conditional on the first of
the SCF's five imputations and on the eligible panel; they omit imputation and survey-design variance, and the
eligible panel excludes households with less than \$5,000 in financial assets. Apart from
one scripted exchange, we test single prompts, not advisory conversations. Real clients may answer a
clarifying question less clearly than our scripts, an adviser may not ask, and a conversation could also widen the
swing if the model anchors on the client's latest statement; the same-person design can test which happens. We also test each model at one point in time and in English only. The
group analysis is observational: the regression separates age from wealth and income, but not from unmeasured
traits correlated with them.

\section*{Ethical considerations statement}
All data are public and de-identified: the Federal Reserve Board's public Survey of Consumer Finances files
\citep{scf,scf2022}, which are in the public domain, and the Basel--Berlin Risk Study data (Frey et al.,
\url{https://osf.io/rce7g}, CC BY 4.0), from which we derived each person's two financial risk ratings and
answers to three investment questions. No new participants were recruited and no individual is identifiable from the profiles, which
use rounded values. Model outputs were collected through OpenRouter. We checked its terms and those of
each model provider as of October 2026: each assigns ownership of outputs to the user or claims none, none
prohibits benchmarking, and their rules on personalised financial advice concern advice delivered to people, which
this audit never did. The released answers are for research and auditing, not financial advice, and
not training data. An adverse use of this work would be to tune an adviser to pass a retest audit on the panel
used here without improving its behaviour on real clients; we therefore release the method, not a fixed test set,
and recommend audits on the deployer's own returning clients.

\section*{Generative AI usage statement}
Generative AI (Claude, Anthropic; Opus 5.5) was used as a coding assistant for the analysis scripts and to draft and edit text. The authors directed the work, made every design decision, and reviewed every analysis and every sentence; all numbers in the paper are produced by the released scripts from the saved model
answers. Claude models are among the systems audited; the main audit's protocol, prompts and sample were fixed before
any results were seen, the clarification test's plan and analysis code were committed before any of its answers
were collected, and every prompt, including those of the later checks, applies identically to every model.

\bibliographystyle{ACM-Reference-Format}
\bibliography{references}

\appendix
\raggedbottom
\section{Models and prompts}\label{app:prompts}

\paragraph{Models and settings.} All requests went through OpenRouter's chat-completions API with its default
provider routing (the serving provider was not recorded), between 9 and 11 October 2026. Identifiers:
\nolinkurl{anthropic/claude-sonnet-5.5}, \nolinkurl{anthropic/claude-haiku-5.5}, \nolinkurl{openai/gpt-6.1-sol},
\nolinkurl{openai/gpt-6-luna}, \nolinkurl{google/gemini-3.8-flash}, \nolinkurl{x-ai/grok-4.7},
\nolinkurl{mistralai/mistral-large-4-0}. Each request carried one user message (plus a system message for the fiduciary
rewording) and no sampling parameters, so temperature, output length and reasoning settings are each provider's
defaults; the check at temperature 0 set only the temperature. The stock share was read from the first
\texttt{"stock\_share"} number in the reply (for the three-asset rewording, stocks as a share of the three numbers).
Every reply is in the released answer files, and the released code rebuilds every request.

\begin{table}[H]
\centering\small
\begin{adjustbox}{max width=\linewidth}\begin{tabular}{lrrrr}
\toprule
Run & Requests & Readable & Retried & Fees (\$) \\
\midrule
Main audit, first 200 households (two answers, two repeats) & 5{,}600 & 5{,}597 & 301 & 10.52 \\
Main audit, other 600 households (two answers) & 8{,}400 & 8{,}400 & 2 & 15.97 \\
Noise note, first 200 SCF households (two answers) & 2{,}800 & 2{,}799 & 0 & 6.97 \\
Basel--Berlin, 171 people (two ratings; rating, note, money questions) & 7{,}182 & 7{,}182 & 0 & 14.92 \\
Basel--Berlin, first run (rating shown one point high; not used) & 7{,}182 & 7{,}182 & 36 & 15.44 \\
Rewordings, first 100 households (three prompts) & 4{,}200 & 4{,}200 & 1 & 9.16 \\
Response curve, 100 SCF 2022 households (answers 0--10) & 7{,}700 & 7{,}700 & 1 & 16.36 \\
Temperature 0, first 100 households & 1{,}400 & 1{,}400 & 0 & 2.56 \\
Reliable-answer note, first 200 SCF households (two answers) & 2{,}800 & 2{,}800 & 0 & 5.81 \\
Clarification test, pilot, 25 other SCF households & 1{,}750 & 1{,}750 & 0 & 3.66 \\
Clarification test, first 200 SCF households (four conditions, two repeats) & 28{,}000 & 28{,}000 & 0 & 59.88 \\
\midrule
Total & 77{,}014 & 77{,}010 & 341 & 161.27 \\
\bottomrule
\end{tabular}
\end{adjustbox}
\caption{Every request behind the paper. A request is one model, profile, prompt and repeat. Readable: replies from
which a stock share could be read. Retried: extra attempts after a rate limit or provider error, or a request sent
twice; fees are those reported for every recorded attempt. Two answers: the 2007 and the 2009 answer.}
\label{tab:accounting}
\end{table}

\paragraph{SCF profile (main audit).}
\begin{quote}\small\ttfamily
You are acting as a financial adviser. A new client has given you the following information.\\[2pt]
Age: 46\\
Education: high school diploma\\
Household income: \$43,000 per year\\
Net worth: \$219,000\\
Financial assets (savings, investments and retirement accounts): \$37,000\\
Risk question: "Which of these statements comes closest to describing the amount of financial risk that you are
willing to take when you save or make investments?"\\
Client's answer: "Take average financial risks expecting to earn average returns."\\[2pt]
Recommend what percentage of the client's financial assets should be held in stocks and stock funds, including
stock held inside retirement accounts. Reply with only a JSON object of the form \{"stock\_share": <number from 0
to 100>\}.
\end{quote}

\paragraph{Noise note (SCF)} Inserted before the request: \emph{``Note: answers to this risk question are not
very reliable. When the same people answer it again a couple of years later, about 4 in 10 choose a different
statement.''}

\paragraph{Reliable-answer note (SCF)} Inserted at the same place: \emph{``Note: answers to this risk question
are very reliable. When the same people answer it again a couple of years later, almost all choose the same
statement.''}

\paragraph{Basel--Berlin profile.} Country, a fixed age band (``late twenties''), education, savings of 20,000 in
local currency, the question ``How would you rate your willingness to take risks in financial matters, from 0 (not
at all willing) to 10 (very willing)?'' and the rating; the note variant says ``about a third move by two points or
more'' after six months; the money variant lists the three investing items with the client's 1--5 likelihood.

\paragraph{Rewordings.} Fiduciary: a system prompt stating that the adviser is a fiduciary who must act in the
client's best interest, and the profile as a list. Client voice: the client describes their own situation in the
first person. Three assets: the same profile, asking for a stocks/bonds/cash split summing to 100. Full text in the
released code (\texttt{llm\_robust.py}).

\section{The benchmark estimate}\label{app:lambda}

Let $m(Z)$ be the best linear prediction of the answer $x$ from the facts $Z$, so that $x - m$ is uncorrelated with
$m$ and $\mathrm{var}(m) = R^2\,\mathrm{var}(x)$. Because the noise $e$ is uncorrelated with $t$ and with $Z$, $m$ is
also the best linear prediction of the attitude $t$ from $Z$, and $t - m$ is uncorrelated with $m$ and with $e$. The
best linear prediction of $t$ from $x$ and $Z$ is then $m + \lambda(x - m)$, with
\begin{equation*}
\lambda = \frac{\mathrm{cov}(t - m,\; x - m)}{\mathrm{var}(x - m)}
        = \frac{\mathrm{var}(t - m)}{\mathrm{var}(x - m)}
        = \frac{\mathrm{var}(t) - \mathrm{var}(m)}{\mathrm{var}(x) - \mathrm{var}(m)}
        = \frac{\rho - R^2}{1 - R^2},
\end{equation*}
where the second step uses $x - m = (t - m) + e$ and the last divides through by $\mathrm{var}(x)$. With no
informative facts ($R^2 = 0$) this is Kelley's estimate, $\lambda = \rho$ \citep{kelley}. $\lambda$ is the
reliability of the part of the answer that the facts do not predict.

\paragraph{On the latent scale.} Treat each four-level answer as a coarse cut of a continuous response
$x^* = t^* + e^*$, with $e^*$ independent of $t^*$ and of $Z$, and with $t^*$ and $e^*$ normal given $Z$. An ordered
probit of the 2007 answer on the facts (panel-weighted, 2,859 households) gives $m^*(Z) = Z\beta$ and, because the
probit fixes $\mathrm{var}(x^* \mid Z) = 1$, $R^{2*} = \mathrm{var}(Z\beta)/(\mathrm{var}(Z\beta) + 1)$. The
weighted polychoric correlation of the 2007 and 2009 answers estimates $\rho^* = \mathrm{var}(t^*)/\mathrm{var}(x^*)$
(two-step, with thresholds from the margins). The derivation above then gives $\lambda^* = (\rho^* - R^{2*})/(1 -
R^{2*})$ on the same scale. The model's input is the discrete answer $k$, so we place it on the latent scale by its
expected value, a truncated-normal mean: with thresholds $\tau_0 = -\infty < \tau_1 < \tau_2 < \tau_3 < \tau_4 =
\infty$ estimated in the same probit and $\phi, \Phi$ the standard normal density and distribution,
\begin{equation*}
s = \mathrm{E}[x^* \mid k, Z] = Z\beta + \frac{\phi(\tau_{k-1} - Z\beta) - \phi(\tau_k - Z\beta)}
{\Phi(\tau_k - Z\beta) - \Phi(\tau_{k-1} - Z\beta)}.
\end{equation*}
We then refit each model's rule as advice on $s$ and the facts, and feed the rule $\mathrm{E}[t^* \mid k, Z] = m^* + \lambda^*(s - m^*)$. Under these assumptions
this step loses nothing: $\mathrm{E}[t^* \mid x^*, Z]$ is linear in $x^*$, so averaging it over the $x^*$ consistent
with answer $k$ gives the same linear function of $s$. (The two-step polychoric estimate treats the margins as
normal, which holds only approximately when $x^*$ is normal given $Z$. The exactness concerns the estimate of
attitude only; the advice rule is still a linear approximation to each model's behaviour, as on the numeric scale.) Estimates: $\rho^* = 0.56$ [0.52, 0.60], $R^{2*} = 0.26$,
$\lambda^* = 0.40$ [0.34, 0.46] (500 household bootstrap draws); excess over the original benchmark 2.5 to 2.8.

\paragraph{Original and corrected benchmarks.} Write a model's fitted rule as advice $= c + b\,x + g^\top Z$. The
original benchmark fits it to the plain-prompt answers, with slope $b_0$; the corrected benchmark fits it to the
answers given when the model is told that the answer is very reliable, with slope $b_1$. Fed $\hat t$, a rule moves
by $\lambda\,b\,(x_2 - x_1)$ when the answer changes, so its benchmark swing is $\lambda\,|b|$ times the mean
answer change. Because the observed swing is close to $|b_0|$ times the mean answer change, the original excess is
close to $1/\lambda$ and the corrected excess close to $k/\lambda$, with $k = b_0/b_1$ the share of the instructed
slope the model keeps without the note. The corrected benchmark is the right one if $b_1$ is the model's
undiscounted response to attitude; if the note makes the model lean harder than that, $k$ is too small and the
excess understated, and if the model discounts internally under both prompts, $b_1$ already holds a discount and
the excess is overstated. Table~\ref{tab:byrho} reports both benchmarks, and their paired difference, at the
estimated reliability, at assumed reliabilities from 0.57 to 0.8 (holding $R^2$ and $m(Z)$ at their estimates) and
on the latent scale (rules refitted on $s$, both fed $m^* + \lambda^*(s - m^*)$). All use the first 200 SCF
households with readable advice for both answers under both prompts (198 for Mistral Large 4). Intervals come from
2{,}000 household bootstrap draws, each paired with its own resample of the 2{,}859 panel households for $\rho$,
$R^2$, $m(Z)$ and the latent parameters; the same draws serve every model, reliability and benchmark, so the
differences are paired.

\begin{table}[H]
\centering\small
\begin{adjustbox}{max width=\linewidth}\begin{tabular}{llllllll}
\toprule
 & \multicolumn{4}{c}{SCF households (first 200)} & \multicolumn{3}{c}{Basel--Berlin} \\
\cmidrule(lr){2-5}\cmidrule(lr){6-8}
 & & & \multicolumn{2}{c}{Benchmark} & & & \\
\cmidrule(lr){4-5}
 & Plain & Note & Original & Corrected & Plain & Note & Benchmark \\
\midrule
Claude Sonnet 5.5 & 0.46 {\scriptsize [0.37, 0.57]} & 0.28 {\scriptsize [0.22, 0.35]} & 0.19 {\scriptsize [0.14, 0.24]} & 0.24 {\scriptsize [0.18, 0.31]} & 0.59 {\scriptsize [0.50, 0.70]} & 0.66 {\scriptsize [0.55, 0.77]} & 0.57 {\scriptsize [0.47, 0.67]} \\
Claude Haiku 5.5 & 0.52 {\scriptsize [0.42, 0.64]} & 0.41 {\scriptsize [0.34, 0.50]} & 0.22 {\scriptsize [0.16, 0.28]} & 0.25 {\scriptsize [0.19, 0.31]} & 0.61 {\scriptsize [0.52, 0.72]} & 0.77 {\scriptsize [0.66, 0.89]} & 0.61 {\scriptsize [0.50, 0.71]} \\
GPT-6.1 Sol & 0.52 {\scriptsize [0.42, 0.65]} & 0.50 {\scriptsize [0.39, 0.62]} & 0.25 {\scriptsize [0.19, 0.32]} & 0.26 {\scriptsize [0.19, 0.33]} & 0.61 {\scriptsize [0.51, 0.71]} & 0.62 {\scriptsize [0.52, 0.73]} & 0.60 {\scriptsize [0.50, 0.71]} \\
GPT-6 Luna & 0.56 {\scriptsize [0.44, 0.68]} & 0.54 {\scriptsize [0.44, 0.66]} & 0.27 {\scriptsize [0.20, 0.34]} & 0.28 {\scriptsize [0.21, 0.35]} & 0.59 {\scriptsize [0.49, 0.70]} & 0.60 {\scriptsize [0.50, 0.71]} & 0.60 {\scriptsize [0.50, 0.71]} \\
Gemini 3.8 Flash & 0.52 {\scriptsize [0.40, 0.65]} & 0.45 {\scriptsize [0.36, 0.56]} & 0.26 {\scriptsize [0.19, 0.33]} & 0.26 {\scriptsize [0.19, 0.34]} & 0.59 {\scriptsize [0.49, 0.70]} & 0.62 {\scriptsize [0.52, 0.73]} & 0.60 {\scriptsize [0.50, 0.71]} \\
Grok 4.7 & 0.55 {\scriptsize [0.44, 0.68]} & 0.46 {\scriptsize [0.38, 0.55]} & 0.27 {\scriptsize [0.20, 0.34]} & 0.27 {\scriptsize [0.21, 0.35]} & 0.64 {\scriptsize [0.54, 0.75]} & 0.66 {\scriptsize [0.57, 0.78]} & 0.60 {\scriptsize [0.50, 0.71]} \\
Mistral Large 4 & 0.56 {\scriptsize [0.46, 0.68]} & 0.40 {\scriptsize [0.33, 0.49]} & 0.25 {\scriptsize [0.19, 0.32]} & 0.26 {\scriptsize [0.19, 0.34]} & 0.65 {\scriptsize [0.55, 0.76]} & 0.66 {\scriptsize [0.56, 0.77]} & 0.60 {\scriptsize [0.50, 0.71]} \\
\bottomrule
\end{tabular}
\end{adjustbox}
\caption{Swing/spread ratio for the plain prompt, the prompt with a note that the answer is noisy, and the
reliability benchmark. SCF: the first 200 households, where both notes were tested; the original benchmark uses
each model's rule fitted to its plain-prompt answers, the corrected benchmark the rule fitted when the model is told
that the answer is very reliable (Section~\ref{sec:bench}). Basel--Berlin: original benchmark only, as the
reliable-answer note was not tested there. 95\% bootstrap intervals.}
\label{tab:ratio}
\end{table}

\begin{table}[H]
\centering\small
\begin{adjustbox}{max width=\linewidth}\begin{tabular}{lllllll}
\toprule
 & \multicolumn{5}{c}{Reliability $\rho$} & Latent \\
\cmidrule(lr){2-6}
 & 0.47 (estimated) & 0.57 & 0.6 & 0.68 & 0.8 & scale \\
\midrule
\multicolumn{7}{l}{\emph{Corrected benchmark (rule fitted when told the answer is very reliable)}} \\
\quad Claude Sonnet 5.5 & 2.44 {\scriptsize [2.08, 3.04]} & 1.77 {\scriptsize [1.59, 1.97]} & 1.63 {\scriptsize [1.47, 1.81]} & 1.36 {\scriptsize [1.23, 1.51]} & 1.08 {\scriptsize [0.98, 1.20]} & 1.98 {\scriptsize [1.70, 2.44]} \\
\quad Claude Haiku 5.5 & 3.04 {\scriptsize [2.54, 3.85]} & 2.20 {\scriptsize [1.94, 2.50]} & 2.03 {\scriptsize [1.80, 2.30]} & 1.69 {\scriptsize [1.50, 1.90]} & 1.35 {\scriptsize [1.20, 1.51]} & 2.48 {\scriptsize [2.09, 3.12]} \\
\quad GPT-6.1 Sol & 3.07 {\scriptsize [2.58, 3.86]} & 2.23 {\scriptsize [1.98, 2.50]} & 2.06 {\scriptsize [1.83, 2.31]} & 1.71 {\scriptsize [1.53, 1.91]} & 1.37 {\scriptsize [1.22, 1.51]} & 2.52 {\scriptsize [2.12, 3.14]} \\
\quad GPT-6 Luna & 3.17 {\scriptsize [2.66, 4.00]} & 2.30 {\scriptsize [2.04, 2.61]} & 2.12 {\scriptsize [1.89, 2.40]} & 1.77 {\scriptsize [1.58, 1.98]} & 1.41 {\scriptsize [1.26, 1.58]} & 2.59 {\scriptsize [2.19, 3.23]} \\
\quad Gemini 3.8 Flash & 3.08 {\scriptsize [2.56, 3.92]} & 2.23 {\scriptsize [1.96, 2.55]} & 2.06 {\scriptsize [1.81, 2.35]} & 1.72 {\scriptsize [1.51, 1.94]} & 1.37 {\scriptsize [1.20, 1.54]} & 2.52 {\scriptsize [2.11, 3.17]} \\
\quad Grok 4.7 & 3.16 {\scriptsize [2.65, 4.01]} & 2.29 {\scriptsize [2.03, 2.60]} & 2.12 {\scriptsize [1.88, 2.40]} & 1.76 {\scriptsize [1.56, 1.98]} & 1.41 {\scriptsize [1.25, 1.57]} & 2.59 {\scriptsize [2.19, 3.24]} \\
\quad Mistral Large 4 & 3.29 {\scriptsize [2.77, 4.15]} & 2.38 {\scriptsize [2.13, 2.70]} & 2.20 {\scriptsize [1.97, 2.49]} & 1.83 {\scriptsize [1.64, 2.05]} & 1.46 {\scriptsize [1.31, 1.63]} & 2.69 {\scriptsize [2.28, 3.36]} \\
\addlinespace
\multicolumn{7}{l}{\emph{Original benchmark (rule fitted to plain-prompt answers)}} \\
\quad Claude Sonnet 5.5 & 3.18 {\scriptsize [2.71, 3.96]} & 2.30 {\scriptsize [2.09, 2.56]} & 2.13 {\scriptsize [1.93, 2.36]} & 1.77 {\scriptsize [1.61, 1.94]} & 1.41 {\scriptsize [1.29, 1.55]} & 2.58 {\scriptsize [2.21, 3.18]} \\
\quad Claude Haiku 5.5 & 3.41 {\scriptsize [2.84, 4.30]} & 2.47 {\scriptsize [2.17, 2.81]} & 2.28 {\scriptsize [2.01, 2.58]} & 1.90 {\scriptsize [1.68, 2.14]} & 1.51 {\scriptsize [1.34, 1.71]} & 2.77 {\scriptsize [2.34, 3.45]} \\
\quad GPT-6.1 Sol & 3.13 {\scriptsize [2.63, 3.93]} & 2.27 {\scriptsize [2.03, 2.55]} & 2.09 {\scriptsize [1.88, 2.35]} & 1.74 {\scriptsize [1.56, 1.94]} & 1.39 {\scriptsize [1.25, 1.54]} & 2.57 {\scriptsize [2.17, 3.19]} \\
\quad GPT-6 Luna & 3.21 {\scriptsize [2.69, 4.04]} & 2.33 {\scriptsize [2.07, 2.63]} & 2.15 {\scriptsize [1.91, 2.42]} & 1.79 {\scriptsize [1.60, 2.01]} & 1.43 {\scriptsize [1.28, 1.59]} & 2.62 {\scriptsize [2.22, 3.27]} \\
\quad Gemini 3.8 Flash & 3.07 {\scriptsize [2.56, 3.91]} & 2.23 {\scriptsize [1.95, 2.54]} & 2.06 {\scriptsize [1.80, 2.34]} & 1.71 {\scriptsize [1.50, 1.93]} & 1.37 {\scriptsize [1.20, 1.53]} & 2.51 {\scriptsize [2.10, 3.16]} \\
\quad Grok 4.7 & 3.24 {\scriptsize [2.72, 4.09]} & 2.34 {\scriptsize [2.09, 2.66]} & 2.17 {\scriptsize [1.94, 2.45]} & 1.80 {\scriptsize [1.61, 2.02]} & 1.44 {\scriptsize [1.29, 1.61]} & 2.64 {\scriptsize [2.23, 3.30]} \\
\quad Mistral Large 4 & 3.43 {\scriptsize [2.88, 4.32]} & 2.49 {\scriptsize [2.23, 2.82]} & 2.30 {\scriptsize [2.06, 2.60]} & 1.91 {\scriptsize [1.72, 2.14]} & 1.53 {\scriptsize [1.37, 1.70]} & 2.81 {\scriptsize [2.38, 3.52]} \\
\addlinespace
\multicolumn{7}{l}{\emph{Corrected minus original, paired}} \\
\quad Claude Sonnet 5.5 & $-$0.74 {\scriptsize [$-$0.94, $-$0.61]} & $-$0.54 {\scriptsize [$-$0.62, $-$0.47]} & $-$0.50 {\scriptsize [$-$0.57, $-$0.43]} & $-$0.41 {\scriptsize [$-$0.47, $-$0.36]} & $-$0.33 {\scriptsize [$-$0.38, $-$0.29]} & $-$0.60 {\scriptsize [$-$0.76, $-$0.50]} \\
\quad Claude Haiku 5.5 & $-$0.37 {\scriptsize [$-$0.55, $-$0.23]} & $-$0.27 {\scriptsize [$-$0.38, $-$0.17]} & $-$0.25 {\scriptsize [$-$0.35, $-$0.16]} & $-$0.21 {\scriptsize [$-$0.29, $-$0.13]} & $-$0.16 {\scriptsize [$-$0.23, $-$0.11]} & $-$0.29 {\scriptsize [$-$0.43, $-$0.19]} \\
\quad GPT-6.1 Sol & $-$0.06 {\scriptsize [$-$0.10, $-$0.02]} & $-$0.04 {\scriptsize [$-$0.07, $-$0.01]} & $-$0.04 {\scriptsize [$-$0.07, $-$0.01]} & $-$0.03 {\scriptsize [$-$0.05, $-$0.01]} & $-$0.03 {\scriptsize [$-$0.04, $-$0.01]} & $-$0.05 {\scriptsize [$-$0.08, $-$0.01]} \\
\quad GPT-6 Luna & $-$0.04 {\scriptsize [$-$0.09, 0.02]} & $-$0.03 {\scriptsize [$-$0.06, 0.01]} & $-$0.02 {\scriptsize [$-$0.06, 0.01]} & $-$0.02 {\scriptsize [$-$0.05, 0.01]} & $-$0.02 {\scriptsize [$-$0.04, 0.01]} & $-$0.03 {\scriptsize [$-$0.07, 0.01]} \\
\quad Gemini 3.8 Flash & 0.01 {\scriptsize [$-$0.03, 0.04]} & 0.01 {\scriptsize [$-$0.02, 0.03]} & 0.01 {\scriptsize [$-$0.02, 0.03]} & 0.01 {\scriptsize [$-$0.01, 0.02]} & 0.00 {\scriptsize [$-$0.01, 0.02]} & 0.01 {\scriptsize [$-$0.02, 0.04]} \\
\quad Grok 4.7 & $-$0.07 {\scriptsize [$-$0.19, 0.05]} & $-$0.05 {\scriptsize [$-$0.13, 0.04]} & $-$0.05 {\scriptsize [$-$0.12, 0.03]} & $-$0.04 {\scriptsize [$-$0.10, 0.03]} & $-$0.03 {\scriptsize [$-$0.08, 0.02]} & $-$0.05 {\scriptsize [$-$0.15, 0.05]} \\
\quad Mistral Large 4 & $-$0.14 {\scriptsize [$-$0.31, $-$0.01]} & $-$0.10 {\scriptsize [$-$0.21, $-$0.01]} & $-$0.10 {\scriptsize [$-$0.20, $-$0.01]} & $-$0.08 {\scriptsize [$-$0.16, $-$0.01]} & $-$0.06 {\scriptsize [$-$0.13, $-$0.01]} & $-$0.12 {\scriptsize [$-$0.25, $-$0.01]} \\
\bottomrule
\end{tabular}
\end{adjustbox}
\caption{Observed swing over the benchmark swing, by benchmark and assumed reliability, on the first 200 SCF
households. At $\rho = 0.47$, the estimate, $\lambda = 0.33$; the latent scale uses $\lambda^* = 0.40$. 95\%
bootstrap intervals; the differences are paired.}
\label{tab:byrho}
\end{table}

\clearpage
\section{Swing by household group}\label{app:who}

Table~\ref{tab:who} gives the numbers behind Figure~\ref{fig:who}, with the share of changed answers among
the audited households and the swing among households whose answer changed.

\begin{table}[H]
\centering\small
\begin{adjustbox}{max width=\linewidth}\begin{tabular}{lrrrlrr}
\toprule
 & & \multicolumn{2}{c}{Answer changed} & & & \\
\cmidrule(lr){3-4}
 & Profiles & All eligible & Audited & Swing, all models & Swing if changed & Range across models \\
\midrule
\multicolumn{7}{l}{\textit{Age (2007)}} \\
\quad under 40 & 190 & 53\% & 50\% & 20.4 {\scriptsize [16.9, 24.2]} & 38.4 & 14.1--22.8 \\
\quad 40 to 59 & 381 & 45\% & 44\% & 13.6 {\scriptsize [11.8, 15.4]} & 29.1 & 10.2--16.1 \\
\quad 60 and over & 229 & 34\% & 31\% & 10.0 {\scriptsize [8.0, 11.9]} & 29.2 & 8.0--11.3 \\
\multicolumn{7}{l}{\textit{Net worth (2007)}} \\
\quad bottom third & 251 & 48\% & 43\% & 17.1 {\scriptsize [14.2, 19.9]} & 37.4 & 12.8--20.2 \\
\quad middle third & 275 & 44\% & 45\% & 14.5 {\scriptsize [12.1, 16.8]} & 30.3 & 10.8--15.9 \\
\quad top third & 274 & 40\% & 36\% & 11.2 {\scriptsize [9.3, 13.2]} & 27.7 & 8.1--13.1 \\
\multicolumn{7}{l}{\textit{Income (2007)}} \\
\quad bottom third & 271 & 39\% & 37\% & 14.6 {\scriptsize [12.2, 17.3]} & 37.1 & 11.3--16.9 \\
\quad middle third & 262 & 47\% & 47\% & 15.5 {\scriptsize [13.1, 18.0]} & 31.1 & 11.4--17.9 \\
\quad top third & 267 & 45\% & 41\% & 12.4 {\scriptsize [10.1, 14.6]} & 27.8 & 8.8--14.1 \\
\multicolumn{7}{l}{\textit{Education (2007)}} \\
\quad no high school diploma & 67 & 37\% & 33\% & 11.5 {\scriptsize [7.2, 15.9]} & 32.9 & 9.0--12.8 \\
\quad high school diploma & 228 & 39\% & 39\% & 15.6 {\scriptsize [12.8, 18.4]} & 37.4 & 11.7--18.0 \\
\quad some college & 147 & 51\% & 49\% & 17.1 {\scriptsize [13.7, 20.7]} & 33.0 & 13.3--20.1 \\
\quad bachelor's degree or higher & 358 & 45\% & 42\% & 12.5 {\scriptsize [10.6, 14.5]} & 27.7 & 8.9--14.3 \\
\bottomrule
\end{tabular}
\end{adjustbox}
\caption{Swing by household group (800 SCF panel households). Answer changed: share of the group whose answer
changed, among all 2,859 eligible panel households (panel-weighted) and among the 800 audited households
(unweighted, since they were drawn with probability proportional to the weight). Swing, all models: each household's swing averaged over the
seven models, with 95\% bootstrap intervals. Swing if changed: the same all-model average, among the group's
households whose answer changed. Range: lowest and highest single-model swing.}
\label{tab:who}
\end{table}

\clearpage
\section{Mitigations}\label{app:fix}

Table~\ref{tab:fixes} gives the results of the noise note and the money questions (Section~\ref{sec:fix}).

\begin{table}[H]
\centering\small
\begin{adjustbox}{max width=\linewidth}\begin{tabular}{lrrllrrrl}
\toprule
 & \multicolumn{4}{c}{SCF households (first 200)} & \multicolumn{4}{c}{Basel--Berlin (171 people)} \\
\cmidrule(lr){2-5}\cmidrule(lr){6-9}
 & \multicolumn{2}{c}{Swing} & Note: $\Delta$ ratio & Note: $\Delta$ mean & \multicolumn{3}{c}{Swing} & Money: $\Delta$ mean \\
 & Plain & Note & & & Rating & Note & Money & \\
\midrule
Claude Sonnet 5.5 & 11.2 & 5.2 & -0.18 {\scriptsize [-0.24, -0.13]} & 6.2 {\scriptsize [4.9, 7.5]} & 11.2 & 9.4 & 9.7 & -11.0 {\scriptsize [-12.4, -9.6]} \\
Claude Haiku 5.5 & 14.3 & 8.0 & -0.11 {\scriptsize [-0.17, -0.06]} & 3.1 {\scriptsize [1.6, 4.5]} & 13.6 & 7.9 & 10.6 & -7.3 {\scriptsize [-8.8, -5.8]} \\
GPT-6.1 Sol & 15.6 & 14.2 & -0.02 {\scriptsize [-0.04, -0.01]} & -1.4 {\scriptsize [-1.8, -1.0]} & 14.2 & 13.2 & 11.1 & -11.0 {\scriptsize [-12.3, -9.7]} \\
GPT-6 Luna & 16.9 & 12.4 & -0.01 {\scriptsize [-0.06, 0.03]} & 4.8 {\scriptsize [3.6, 6.0]} & 13.9 & 12.0 & 8.5 & -12.1 {\scriptsize [-14.1, -10.2]} \\
Gemini 3.8 Flash & 16.4 & 12.4 & -0.07 {\scriptsize [-0.10, -0.03]} & 3.5 {\scriptsize [2.4, 4.6]} & 14.2 & 14.1 & 12.7 & -9.1 {\scriptsize [-10.6, -7.6]} \\
Grok 4.7 & 16.9 & 9.7 & -0.09 {\scriptsize [-0.19, 0.00]} & 8.8 {\scriptsize [6.7, 11.0]} & 15.7 & 13.2 & 13.4 & -7.4 {\scriptsize [-8.8, -6.0]} \\
Mistral Large 4 & 18.1 & 9.4 & -0.17 {\scriptsize [-0.26, -0.08]} & 7.7 {\scriptsize [5.3, 10.3]} & 17.7 & 15.1 & 14.8 & -5.8 {\scriptsize [-6.8, -4.7]} \\
\bottomrule
\end{tabular}
\end{adjustbox}
\caption{Two mitigations. SCF (first 200 households, first repeat): swing with the plain prompt and with a note that
the answer is not very reliable, and the note's effect on the swing/spread ratio and on mean advice. Basel--Berlin:
swing with the rating alone, with the note, and with three money questions, and the money questions' effect on
mean advice. Percentage points; 95\% bootstrap intervals for differences.}
\label{tab:fixes}
\end{table}

\paragraph{The clarification test.} The adviser's question is the same in every scenario: ``Which risk statement best describes your current
willingness to take financial risk? If it differs from your previous answer, does that reflect a deliberate change
in preference, a misunderstanding of the question, or uncertainty?'' In scenario B the client replies, for
example, ``I misunderstood the statements. `[previous statement]' still describes my willingness to take risk; my
preference has not changed.'' In C the reply is, for example, ``I have reconsidered my willingness to take risk.
`[today's statement]' now describes my preference; this is a deliberate change.'' In A the client confirms the
previous statement. Each reply has two phrasings, alternating within each direction of change; all are in the
released code. Households whose previous answer was the most or least cautious can move only one way; the rest
were assigned at random so that 100 households move towards more risk and 100 towards less. In the direct and note conditions, B and C are the same
prompt, asked once and used for both. Requests for all conditions were sent in random order, each as a fresh conversation, at default sampling. All 28{,}000 replies were
readable, so no household was dropped. Asking the same prompt twice moved advice by 1.1 to 4.3 points on average,
depending on the model. An interim look at 14\% of the answers, which the plan's log
records, changed no prompt, parsing or analysis step. Averaged over the models, the exchange steadied advice in every age group (swing under 40:
41.9 points at once, 2.5 with the exchange; 60 and over: 28.6 and 1.2) and for the 58 clients whose previous
answer was that they would take no financial risk (40.2 and 0.8). After a confirmed change, advice stayed put or
moved the wrong way for 2\% of households with the exchange and 4\% at once.

\begin{table}[H]
\centering\small
\begin{adjustbox}{max width=\linewidth}\begin{tabular}{lrrrrrrrrlll}
\toprule
 & \multicolumn{4}{c}{Unwanted swing $S$} & \multicolumn{4}{c}{Response to confirmed change $R$} & \multicolumn{2}{c}{Retention of $R$} & Clarify: shift, \\
\cmidrule(lr){2-5}\cmidrule(lr){6-9}\cmidrule(lr){10-11}
 & Direct & Note & Clarify & One turn & Direct & Note & Clarify & One turn & Note & Clarify & stable client \\
\midrule
Claude Sonnet 5.5 & 22.9 & 9.1 & 2.6 & 2.6 & 22.9 & 8.8 & 32.4 & 31.4 & 0.38 {\scriptsize [0.35, 0.41]} & 1.42 {\scriptsize [1.37, 1.47]} & -2.8 {\scriptsize [-3.5, -2.1]} \\
Claude Haiku 5.5 & 30.9 & 11.4 & 2.1 & 2.7 & 30.7 & 11.2 & 35.8 & 33.7 & 0.36 {\scriptsize [0.34, 0.39]} & 1.17 {\scriptsize [1.13, 1.20]} & 0.4 {\scriptsize [-0.2, 1.0]} \\
GPT-6.1 Sol & 33.5 & 29.5 & 1.2 & 0.9 & 33.4 & 29.5 & 35.1 & 35.1 & 0.88 {\scriptsize [0.86, 0.91]} & 1.05 {\scriptsize [1.04, 1.07]} & -0.2 {\scriptsize [-0.5, 0.1]} \\
GPT-6 Luna & 36.0 & 20.2 & 1.5 & 1.4 & 35.5 & 20.0 & 36.2 & 36.3 & 0.56 {\scriptsize [0.53, 0.59]} & 1.02 {\scriptsize [1.00, 1.04]} & -1.1 {\scriptsize [-1.5, -0.7]} \\
Gemini 3.8 Flash & 38.1 & 24.0 & 1.1 & 1.0 & 38.1 & 24.0 & 38.9 & 38.6 & 0.63 {\scriptsize [0.59, 0.66]} & 1.02 {\scriptsize [1.01, 1.04]} & -0.5 {\scriptsize [-0.7, -0.2]} \\
Grok 4.7 & 37.2 & 16.1 & 1.8 & 1.6 & 36.5 & 15.4 & 37.8 & 38.2 & 0.42 {\scriptsize [0.37, 0.47]} & 1.04 {\scriptsize [1.02, 1.06]} & -0.2 {\scriptsize [-0.7, 0.2]} \\
Mistral Large 4 & 36.0 & 14.1 & 2.9 & 4.0 & 35.6 & 13.5 & 41.1 & 40.2 & 0.38 {\scriptsize [0.35, 0.42]} & 1.15 {\scriptsize [1.12, 1.19]} & -0.5 {\scriptsize [-1.1, 0.1]} \\
\midrule
Mean of models & 33.5 & 17.8 & 1.9 & 2.0 & 33.2 & 17.5 & 36.8 & 36.2 & 0.52 {\scriptsize [0.50, 0.54]} & 1.12 {\scriptsize [1.11, 1.14]} & -0.7 {\scriptsize [-0.9, -0.5]} \\
\bottomrule
\end{tabular}
\end{adjustbox}
\caption{Clarification test by model (200 SCF households, two repeats; percentage points). Unwanted swing $S$ and
response to a confirmed change $R$ in the four conditions (one turn: the clarifying question and the client's
reply inside a one-turn prompt); retention of $R$ relative to the direct condition, with 95\% bootstrap intervals;
and the clarifying question's shift of advice for the stable client, with the 90\% interval of the pre-set test.}
\label{tab:clarify}
\end{table}

\end{document}